\documentclass[11pt]{article}
\usepackage{acl2016}
\usepackage{url}
\usepackage{times}
\usepackage{latexsym}
\aclfinalcopy

\usepackage[utf8]{inputenc}
\usepackage[T1]{fontenc}

\usepackage{CJKutf8}

\usepackage[russian,english]{babel}

\usepackage{enumitem}
\usepackage{amsmath}
\usepackage{multirow}
\usepackage{tikz-qtree}
\usepackage{tikz-dependency}
\usepackage{adjustbox}

\usepackage{pgfplots}
\pgfplotsset{compat=newest}

\usepackage{siunitx}
\newcommand{\pipe}{{\color{gray}|}}

\usepackage{textcomp}
\usepackage{algorithm}
\usepackage{listings}{
\lstset{
  language=Python,
  upquote=true,
  showstringspaces=false,
  formfeed=newpage,
  tabsize=4,
  commentstyle=itshape,
  basicstyle=\scriptsize\ttfamily,
  morekeywords={models, lambda, forms}
}

\setlength\titlebox{5cm}

\renewcommand\footnotemark{}
\title{Neural Machine Translation of Rare Words with Subword Units\thanks{The research presented in this publication was conducted in cooperation with Samsung Electronics Polska sp.\ z o.o.\ - Samsung R\&D Institute Poland.} }

\author{anonymous for review}

\author{
Rico Sennrich\and Barry Haddow \and Alexandra Birch\\
School of Informatics, University of Edinburgh\\
{\tt \{rico.sennrich,a.birch\}@ed.ac.uk}, {\tt bhaddow@inf.ed.ac.uk}
}

\date{}

\begin{document}
\maketitle
\begin{abstract}
Neural machine translation (NMT) models typically operate with a fixed vocabulary, but translation is an open-vocabulary problem.
Previous work addresses the translation of out-of-vocabulary words by backing off to a dictionary.
In this paper, we introduce a simpler and more effective approach, making the NMT model capable of open-vocabulary translation by encoding rare and unknown words as sequences of subword units.
This is based on the intuition that various word classes are translatable via smaller units than words, for instance
names (via character copying or transliteration), compounds (via compositional translation), and cognates and loanwords (via phonological and morphological transformations).
We discuss the suitability of different word segmentation techniques, including simple character $n$-gram models and a segmentation based on the \emph{byte pair encoding} compression algorithm,
and empirically show that subword models improve over a back-off dictionary baseline for the \mbox{WMT 15} translation tasks English$\to$German and English$\to$Russian by up to 1.1 and 1.3 {\sc Bleu}, respectively.
\end{abstract}

\section{Introduction}

Neural machine translation has recently shown impressive results \cite{kalchbrenner13emnlp,DBLP:conf/nips/SutskeverVL14,DBLP:journals/corr/BahdanauCB14}.
However, the translation of rare words is an open problem.
The vocabulary of neural models is typically limited to \num{30000}--\num{50000} words, but translation is an open-vocabulary problem,
and especially for languages with productive word formation processes such as agglutination and compounding, translation models require mechanisms that go below the word level.
As an example, consider compounds such as the German \emph{Abwasser\pipe behandlungs\pipe anlange} `sewage water treatment plant', for which a segmented, variable-length representation is intuitively more appealing than encoding the word as a fixed-length vector.

For word-level NMT models, the translation of out-of-vocabulary words has been addressed through a back-off to a dictionary look-up \cite{jean15,DBLP:journals/corr/LuongSLVZ14}.
We note that such techniques make assumptions that often do not hold true in practice.
For instance, there is not always a 1-to-1 correspondence between source and target words because of variance in the degree of morphological synthesis between languages, like in our introductory compounding example.
Also, word-level models are unable to translate or generate unseen words.
Copying unknown words into the target text, as done by \cite{jean15,DBLP:journals/corr/LuongSLVZ14}, is a reasonable strategy for names, but morphological changes and transliteration is often required, especially if alphabets differ.

We investigate NMT models that operate on the level of subword units.
Our main goal is to model open-vocabulary translation in the NMT network itself, without requiring a back-off model for rare words.
In addition to making the translation process simpler, we also find that the subword models achieve better accuracy for the translation of rare words than large-vocabulary models and back-off dictionaries, 
and are able to productively generate new words that were not seen at training time.
Our analysis shows that the neural networks are able to learn compounding and transliteration from subword representations.

This paper has two main contributions:

\begin{itemize}
\setlength{\itemsep}{0pt}
\item We show that open-vocabulary neural machine translation is possible by encoding (rare) words via subword units.
We find our architecture simpler and more effective than using large vocabularies and back-off dictionaries \cite{jean15,DBLP:journals/corr/LuongSLVZ14}.

\item We adapt \emph{byte pair encoding} (BPE) \cite{Gage:1994:NAD:177910.177914}, a compression algorithm, to the task of word segmentation.
BPE allows for the representation of an open vocabulary through a fixed-size vocabulary of variable-length character sequences,
making it a very suitable word segmentation strategy for neural network models.

\end{itemize}

\section{Neural Machine Translation}

We follow the neural machine translation architecture by \newcite{DBLP:journals/corr/BahdanauCB14}, which we will briefly summarize here.
However, we note that our approach is not specific to this architecture.

The neural machine translation system is implemented as an encoder-decoder network with recurrent neural networks.

The encoder is a bidirectional neural network with gated recurrent units \cite{cho-EtAl:2014:EMNLP2014} that reads an input sequence $x=(x_1,...,x_m)$ and calculates a forward sequence of hidden states $(\overrightarrow{h}_1,...,\overrightarrow{h}_m)$,
and a backward sequence $(\overleftarrow{h}_1,...,\overleftarrow{h}_m)$.
The hidden states $\overrightarrow{h}_j$ and $\overleftarrow{h}_j$ are concatenated to obtain the annotation vector $h_j$.

The decoder is a recurrent neural network that predicts a target sequence $y=(y_1,...,y_n)$.
Each word $y_i$ is predicted based on a recurrent hidden state $s_i$, the previously predicted word $y_{i-1}$, and a context vector $c_i$.
$c_i$ is computed as a weighted sum of the annotations $h_j$.
The weight of each annotation $h_j$ is computed through an \emph{alignment model} $\alpha_{ij}$, which models the probability that $y_i$ is aligned to $x_j$.
The alignment model is a single-layer feedforward neural network that is learned jointly with the rest of the network through backpropagation.

A detailed description can be found in \cite{DBLP:journals/corr/BahdanauCB14}.
Training is performed on a parallel corpus with stochastic gradient descent.
For translation, a beam search with small beam size is employed.

\section{Subword Translation}

The main motivation behind this paper is that the translation of some words is transparent in that they are translatable by a competent translator even if they are novel to him or her, based on a translation of known subword units such as morphemes or phonemes.
Word categories whose translation is potentially transparent include:

\begin{itemize}
\setlength{\itemsep}{0pt}
\item named entities. Between languages that share an alphabet, names can often be copied from source to target text.
Transcription or transliteration may be required, especially if the alphabets or syllabaries differ. Example:\\
Barack Obama (English; German)\\
\foreignlanguage{russian}{Барак Обама} (Russian)\\
\begin{CJK*}{UTF8}{goth}
\small
\CJKnospace
\ignorespaces
バラク・オバマ
\end{CJK*}
(ba-ra-ku o-ba-ma) (Japanese)
\item cognates and loanwords. Cognates and loanwords with a common origin can differ in regular ways between languages, so that character-level translation rules are sufficient \cite{tiedemann:2012:EACL2012}. Example:\\
claustrophobia (English)\\
Klaustrophobie (German)\\
\foreignlanguage{russian}{Клаустрофобия} (Klaustrofobiâ) (Russian)
\item morphologically complex words. Words containing multiple morphemes, for instance formed via compounding, affixation, or inflection, may be translatable by translating the morphemes separately. Example:\\
solar system (English)\\
Sonnensystem (Sonne + System) (German)\\
Naprendszer (Nap + Rendszer) (Hungarian)
\end{itemize}

In an analysis of 100 rare tokens (not among the \num{50000} most frequent types) in our German training data\footnote{Primarily parliamentary proceedings and web crawl data.}, the majority of tokens are potentially translatable from English through smaller units.
We find 56 compounds, 21 names, 6 loanwords with a common origin (\emph{emancipate}$\to$\emph{emanzipieren}),
5 cases of transparent affixation (\emph{sweetish} `sweet' + `-ish' $\to$ \emph{süßlich} `süß' + `-lich'), 1 number and 1 computer language identifier.

Our hypothesis is that a segmentation of rare words into appropriate subword units is sufficient to allow for the neural translation network to learn transparent translations, and to generalize this knowledge to translate and produce unseen words.\footnote{Not every segmentation we produce is transparent. While we expect no performance benefit from opaque segmentations, i.e.\ segmentations where the units cannot be translated independently, our NMT models show robustness towards oversplitting.}
We provide empirical support for this hypothesis in Sections \ref{sec-eval} and \ref{sec-analysis}.
First, we discuss different subword representations.

\subsection{Related Work}
\label{related-sec}

For Statistical Machine Translation (SMT), the translation of unknown words has been the subject of intensive research.

A large proportion of unknown words are names, which can just be copied into the target text if both languages share an alphabet.
If alphabets differ, transliteration is required \cite{DBLP:conf/eacl/DurraniSHK14}.
Character-based translation has also been investigated with phrase-based models, which proved especially successful for closely related languages \cite{vilar-peter-ney:2007:WMT,Tiedemann474916,DBLP:conf/acl/NeubigWMK12}.

The segmentation of morphologically complex words such as compounds is widely used for SMT, and various algorithms for morpheme segmentation have been investigated \cite{Niessen00improvingsmt,koehn03b,virpioja07mtsummit,DBLP:conf/acl/StallardDKLB12}.
Segmentation algorithms commonly used for phrase-based SMT tend to be conservative in their splitting decisions,
whereas we aim for an aggressive segmentation that allows for open-vocabulary translation with a compact network vocabulary, and without having to resort to back-off dictionaries.

The best choice of subword units may be task-specific.
For speech recognition, phone-level language models have been used \cite{DBLP:conf/interspeech/BazziG00}.
\newcite{mikolov2012} investigate subword language models, and propose to use syllables.
For multilingual segmentation tasks, multilingual algorithms have been proposed \cite{snyder-barzilay:2008:ACLMain}.
We find these intriguing, but inapplicable at test time.

Various techniques have been proposed to produce fixed-length continuous word vectors based on characters or morphemes \cite{DBLP:conf/conll/LuongSM13,Botha2014,ling15,DBLP:journals/corr/KimJSR15}.
An effort to apply such techniques to NMT, parallel to ours, has found no significant improvement over word-based approaches \cite{2015arXiv151104586L}.
One technical difference from our work is that the attention mechanism still operates on the level of words in the model by \newcite{2015arXiv151104586L}, and that the representation of each word is fixed-length.
We expect that the attention mechanism benefits from our variable-length representation: the network can learn to place attention on different subword units at each step.
Recall our introductory example \emph{Abwasserbehandlungsanlange}, for which a subword segmentation avoids the information bottleneck of a fixed-length representation.

Neural machine translation differs from phrase-based methods in that there are strong incentives to minimize the vocabulary size of neural models to increase time and space efficiency, and to allow for translation without back-off models.
At the same time, we also want a compact representation of the text itself, since an increase in text length reduces efficiency and increases the distances over which neural models need to pass information.

A simple method to manipulate the trade-off between vocabulary size and text size is to use shortlists of unsegmented words, using subword units only for rare words.
As an alternative, we propose a segmentation algorithm based on byte pair encoding (BPE), which lets us learn a vocabulary that provides a good compression rate of the text.

\subsection{Byte Pair Encoding (BPE)}

Byte Pair Encoding (BPE) \cite{Gage:1994:NAD:177910.177914} is a simple data compression technique that iteratively replaces the most frequent pair of bytes in a sequence with a single, unused byte.
We adapt this algorithm for word segmentation.
Instead of merging frequent pairs of bytes, we merge characters or character sequences.

Firstly, we initialize the symbol vocabulary with the character vocabulary, and represent each word as a sequence of characters, plus a special end-of-word symbol `$\cdot$', which allows us to restore the original tokenization after translation.
We iteratively count all symbol pairs and replace each occurrence of the most frequent pair (`A', `B') with a new symbol `AB'.
Each merge operation produces a new symbol which represents a character $n$-gram.
Frequent character $n$-grams (or whole words) are eventually merged into a single symbol, thus BPE requires no shortlist.
The final symbol vocabulary size is equal to the size of the initial vocabulary, plus the number of merge operations -- the latter is the only hyperparameter of the algorithm.

\begin{algorithm}
\begin{lstlisting}
import re, collections

def get_stats(vocab):
  pairs = collections.defaultdict(int)
  for word, freq in vocab.items():
    symbols = word.split()
    for i in range(len(symbols)-1):
      pairs[symbols[i],symbols[i+1]] += freq
  return pairs

def merge_vocab(pair, v_in):
  v_out = {}
  bigram = re.escape(' '.join(pair))
  p = re.compile(r'(?<!\S)' + bigram + r'(?!\S)')
  for word in v_in:
    w_out = p.sub(''.join(pair), word)
    v_out[w_out] = v_in[word]
  return v_out

vocab = {'l o w </w>' : 5, 'l o w e r </w>' : 2,
         'n e w e s t </w>':6, 'w i d e s t </w>':3}
num_merges = 10
for i in range(num_merges):
  pairs = get_stats(vocab)
  best = max(pairs, key=pairs.get)
  vocab = merge_vocab(best, vocab)
  print(best)
\end{lstlisting}
\caption{Learn BPE operations}
\label{bpe-algorithm}
\end{algorithm}

For efficiency, we do not consider pairs that cross word boundaries.
The algorithm can thus be run on the dictionary extracted from a text, with each word being weighted by its frequency.
A minimal Python implementation is shown in Algorithm \ref{bpe-algorithm}.
In practice, we increase efficiency by indexing all pairs, and updating data structures incrementally.

The main difference to other compression algorithms, such as Huffman encoding, which have been proposed to produce a variable-length encoding of words for NMT \cite{chitnis15},
is that our symbol sequences are still interpretable as subword units, and that the network can generalize to translate and produce new words (unseen at training time) on the basis of these subword units.

\begin{figure}
\centering
\small{
\begin{tabular}{lcl}
r $\cdot$ & $\rightarrow$ & r$\cdot$ \\
l o & $\rightarrow$ & lo\\
lo w & $\rightarrow$ & low\\
e r$\cdot$ & $\rightarrow$ & er$\cdot$ \\
\end{tabular}}
\caption{BPE merge operations learned from dictionary \{`low', `lowest', `newer', `wider'\}.}
\label{bpe-example}
\end{figure}

Figure \ref{bpe-example} shows a toy example of learned BPE operations.
At test time, we first split words into sequences of characters, then apply the learned operations to merge the characters into larger, known symbols.
This is applicable to any word, and allows for open-vocabulary networks with fixed symbol vocabularies.\footnote{The only symbols that will be unknown at test time are unknown characters, or symbols of which all occurrences in the training text have been merged into larger symbols, like `safeguar', which has all occurrences in our training text merged into `safeguard'. We observed no such symbols at test time, but the issue could be easily solved by recursively reversing specific merges until all symbols are known.}
In our example, the OOV `lower' would be segmented into `low er$\cdot$'.

We evaluate two methods of applying BPE:
learning two independent encodings, one for the source, one for the target vocabulary, or learning the encoding on the union of the two vocabularies (which we call \emph{joint BPE}).\footnote{In practice, we simply concatenate the source and target side of the training set to learn joint BPE.}
The former has the advantage of being more compact in terms of text and vocabulary size, and having stronger guarantees that each subword unit has been seen in the training text of the respective language,
whereas the latter improves consistency between the source and the target segmentation.
If we apply BPE independently, the same name may be segmented differently in the two languages, which makes it harder for the neural models to learn a mapping between the subword units.
To increase the consistency between English and Russian segmentation despite the differing alphabets, we transliterate the Russian vocabulary into Latin characters with ISO-9 to learn the joint BPE encoding,
then transliterate the BPE merge operations back into Cyrillic to apply them to the Russian training text.\footnote{Since the Russian training text also contains words that use the Latin alphabet, we also apply the Latin BPE operations.}

\section{Evaluation}
\label{sec-eval}

We aim to answer the following empirical questions:

\begin{itemize}
\setlength{\itemsep}{0pt}
\item Can we improve the translation of rare and unseen words in neural machine translation by representing them via subword units?
\item Which segmentation into subword units performs best in terms of vocabulary size, text size, and translation quality?
\end{itemize}

We perform experiments on data from the shared translation task of WMT 2015.
For English$\to$German, our training set consists of 4.2 million sentence pairs, or approximately 100 million tokens.
For English$\to$Russian, the training set consists of 2.6 million sentence pairs, or approximately 50 million tokens.
We tokenize and truecase the data with the scripts provided in Moses \cite{koehnmoses}.
We use newstest2013 as development set, and report results on newstest2014 and newstest2015.

We report results with {\sc Bleu} (\emph{mteval-v13a.pl}),
and {\sc chrF3} \cite{popovic:2015:WMT}, a character n-gram F$_3$ score which was found to correlate well with human judgments, especially for translations out of English \cite{stanojevic-EtAl:2015:WMT}. 
Since our main claim is concerned with the translation of rare and unseen words, we report separate statistics for these.
We measure these through unigram F$_1$, which we calculate as the harmonic mean of clipped unigram precision and recall.\footnote{Clipped unigram precision is essentially 1-gram BLEU without brevity penalty.}

We perform all experiments with Groundhog\footnote{\url{github.com/sebastien-j/LV_groundhog}} \cite{DBLP:journals/corr/BahdanauCB14}.
We generally follow settings by previous work \cite{DBLP:journals/corr/BahdanauCB14,jean15}.
All networks have a hidden layer size of 1000, and an embedding layer size of 620.
Following \newcite{jean15}, we only keep a shortlist of $\tau=30000$ words in memory.

During training, we use Adadelta \cite{DBLP:journals/corr/abs-1212-5701}, a minibatch size of 80, and reshuffle the training set between epochs.
We train a network for approximately 7 days, then take the last 4 saved models (models being saved every 12 hours), and continue training each with a fixed embedding layer (as suggested by \cite{jean15}) for 12 hours.
We perform two independent training runs for each models, once with cut-off for gradient clipping \cite{DBLP:conf/icml/PascanuMB13} of 5.0, once with a cut-off of 1.0 -- the latter produced better single models for most settings.
We report results of the system that performed best on our development set (newstest2013), and of an ensemble of all 8 models.

We use a beam size of 12 for beam search, with probabilities normalized by sentence length.
We use a bilingual dictionary based on fast-align \cite{dyer-chahuneau-smith:2013:NAACL-HLT}.
For our baseline, this serves as back-off dictionary for rare words.
We also use the dictionary to speed up translation for all experiments, only performing the softmax over a filtered list of candidate translations (like \newcite{jean15}, we use $K=30000$; $K'=10$).

\subsection{Subword statistics}

Apart from translation quality, which we will verify empirically, our main objective is to represent an open vocabulary through a compact fixed-size subword vocabulary, and allow for efficient training and decoding.\footnote{The time complexity of encoder-decoder architectures is at least linear to sequence length, and oversplitting harms efficiency.}

Statistics for different segmentations of the German side of the parallel data are shown in Table \ref{splitting}.
A simple baseline is the segmentation of words into character $n$-grams.\footnote{Our character n-grams do not cross word boundaries. We mark whether a subword is word-final or not with a special character, which allows us to restore the original tokenization.}
Character $n$-grams allow for different trade-offs between sequence length (\# tokens) and vocabulary size (\# types), depending on the choice of $n$.
The increase in sequence length is substantial; one way to reduce sequence length is to leave a shortlist of the $k$ most frequent word types unsegmented.
Only the unigram representation is truly open-vocabulary.
However, the unigram representation performed poorly in preliminary experiments, and we report translation results with a bigram representation, which is empirically better, but unable to produce some tokens in the test set with the training set vocabulary.

We report statistics for several word segmentation techniques that have proven useful in previous SMT research, including
frequency-based compound splitting \cite{koehn03b}, rule-based hyphenation \cite{Liang:151530}, and Morfessor \cite{creutz-lagus:2002:ACL02-MPL}.
We find that they only moderately reduce vocabulary size, and do not solve the unknown word problem, and we thus find them unsuitable for our goal of open-vocabulary translation without back-off dictionary.

BPE meets our goal of being open-vocabulary, and the learned merge operations can be applied to the test set to obtain a segmentation with no unknown symbols.\footnote{Joint BPE can produce segments that are unknown because they only occur in the English training text, but these are rare (0.05\% of test tokens).}
Its main difference from the character-level model is that the more compact representation of BPE allows for shorter sequences, and that the attention model operates on variable-length units.\footnote{We highlighted the limitations of word-level attention in section \ref{related-sec}. At the other end of the spectrum, the character level is suboptimal for alignment \cite{Tiedemann474916}.}
Table \ref{splitting} shows BPE with \num{59500} merge operations, and joint BPE with \num{89500} operations.

In practice, we did not include infrequent subword units in the NMT network vocabulary, since there is noise in the subword symbol sets, e.g.\ because of characters from foreign alphabets.
Hence, our network vocabularies in Table \ref{results} are typically slightly smaller than the number of types in Table \ref{splitting}.

\begin{table}
\centering
\small{
\begin{tabular}{lrrrr}
segmentation & \# tokens & \# types &\# UNK \\
\hline
none & 100 m & \num{1750000} & 1079 \\
characters & 550 m & \num{3000} & 0\\
character bigrams & 306 m & \num{20000} & 34\\
character trigrams & 214 m & \num{120000} & 59 \\
\hline
compound splitting$^\triangle$ & 102 m & \num{1100000} & 643 \\
morfessor* & 109 m & \num{544000} & 237\\
hyphenation$^\diamond$ & 186 m & \num{404000} & 230\\
\hline
BPE & 112 m & \num{63000} & 0 \\
BPE (joint) & 111 m & \num{82000} & 32 \\
\hline
character bigrams & \multirow{2}{*}{129 m} & \multirow{2}{*}{\num{69000}} & \multirow{2}{*}{34} \\
(shortlist: \num{50000}) \\
\end{tabular}}
\caption{Corpus statistics for German training corpus with different word segmentation techniques. \#UNK: number of unknown tokens in newstest2013. $\triangle$: \cite{koehn03b}; *: \cite{creutz-lagus:2002:ACL02-MPL}; $\diamond$: \cite{Liang:151530}.}
\label{splitting}
\end{table}

\subsection{Translation experiments}

\begin{table*}
\centering
\small
\setlength{\tabcolsep}{3pt}
\begin{tabular}{llrrr|rr|rr|rrr}
&&&\multicolumn{2}{c|}{vocabulary}&\multicolumn{2}{c|}{{\sc Bleu}} & \multicolumn{2}{c|}{{\sc chrF3}} & \multicolumn{3}{c}{unigram F$_1$ (\%)}\\
name & segmentation & shortlist & source & target & single & ens-8 & single & ens-8 & all & rare & OOV\\
\hline
\multicolumn{4}{l}{syntax-based \cite{sennrichhaddow15}} & & 
24.4 & - & 
55.3 & - & 59.1 & 46.0 & 37.7\\ 
\hline
WUnk & - & - & \num{300000} & \num{500000} & 
20.6 & 22.8 & 
47.2 & 48.9 & 56.7 & 20.4 & 0.0\\ 
WDict & - & - & \num{300000} & \num{500000} & 
22.0 & 24.2 & 
50.5 & 52.4 & 58.1 & 36.8 & \textbf{36.8}\\ 
C2-50k &char-bigram & \num{50000} & \num{60000} & \num{60000} & 
\textbf{22.8} & \textbf{25.3} & 
51.9 & 53.5 & 58.4 & 40.5 & 30.9 \\ 
BPE-60k &BPE & - & \num{60000} & \num{60000} & 
21.5 & 24.5 & 
\textbf{52.0} & 53.9 & 58.4 & 40.9 & 29.3\\ 
BPE-J90k &BPE (joint) & - & \num{90000} & \num{90000} & 
\textbf{22.8} & 24.7 & 
51.7 & \textbf{54.1} & \textbf{58.5} & \textbf{41.8} & 33.6\\ 
\end{tabular}
\caption{English$\to$German translation performance ({\sc Bleu}, {\sc chrF3} and unigram F$_1$) on newstest2015.
Ens-8: ensemble of 8 models.
Best NMT system in bold.
Unigram F$_1$ (with ensembles) is computed for all words ($n=44085$), rare words (not among top \num{50000} in training set; $n=2900$), and OOVs (not in training set; $n=1168$).}
\label{results}
\end{table*}

English$\to$German translation results are shown in Table \ref{results};
English$\to$Russian results in Table \ref{results-russian}.

Our baseline \textbf{WDict} is a word-level model with a back-off dictionary.
It differs from \textbf{WUnk} in that the latter uses no back-off dictionary, and just represents out-of-vocabulary words as UNK\footnote{We use \emph{UNK} for words that are outside the model vocabulary, and \emph{OOV} for those that do not occur in the training text.}.
The back-off dictionary improves unigram F$_1$ for rare and unseen words, although the improvement is smaller for English$\to$Russian, since the back-off dictionary is incapable of transliterating names.

All subword systems operate without a back-off dictionary.
We first focus on unigram F$_1$, where all systems improve over the baseline,
especially for rare words (36.8\%$\rightarrow$41.8\% for EN$\to$DE; 26.5\%$\rightarrow$29.7\% for EN$\to$RU).
For OOVs, the baseline strategy of copying unknown words works well for English$\to$German.
However, when alphabets differ, like in English$\to$Russian, the subword models do much better.

Unigram F$_1$ scores indicate that learning the BPE symbols on the vocabulary union (\textbf{BPE-J90k})
is more effective than learning them separately (\textbf{BPE-60k}), and more effective than using character bigrams with a shortlist of \num{50000} unsegmented words (\textbf{C2-50k}),
but all reported subword segmentations are viable choices and outperform the back-off dictionary baseline.

Our subword representations cause big improvements in the translation of rare and unseen words, but these only constitute 9-11\% of the test sets.
Since rare words tend to carry central information in a sentence, we suspect that {\sc Bleu} and {\sc chrF3} underestimate their effect on translation quality.
Still, we also see improvements over the baseline in total unigram F$_1$, as well as {\sc Bleu} and {\sc chrF3}, and the subword ensembles outperform the WDict baseline by 0.3--1.3 {\sc Bleu} and 0.6--2 {\sc chrF3}.
There is some inconsistency between {\sc Bleu} and {\sc chrF3}, which we attribute to the fact that {\sc Bleu} has a precision bias, and {\sc chrF3} a recall bias.

For English$\to$German, we observe the best {\sc Bleu} score of 25.3 with C2-50k, but the best {\sc chrF3} score of 54.1 with BPE-J90k.
For comparison to the (to our knowledge) best non-neural MT system on this data set, we report syntax-based SMT results \cite{sennrichhaddow15}.
We observe that our best systems outperform the syntax-based system in terms of {\sc Bleu}, but not in terms of {\sc chrF3}.
Regarding other neural systems, \newcite{luong-pham-manning:2015:EMNLP} report a {\sc Bleu} score of 25.9 on newstest2015, but we note that they use an ensemble of 8 independently trained models, and also report strong improvements from applying dropout, which we did not use.
We are confident that our improvements to the translation of rare words are orthogonal to improvements achievable through other improvements in the network architecture, training algorithm, or better ensembles.

For English$\to$Russian, the state of the art is the phrase-based system by \newcite{haddow-EtAl:2015:WMT}.
It outperforms our WDict baseline by 1.5 {\sc Bleu}.
The subword models are a step towards closing this gap, and BPE-J90k yields an improvement of 1.3 {\sc Bleu}, and 2.0 {\sc chrF3}, over WDict.

\begin{table*}
\centering
\small
\setlength{\tabcolsep}{3pt}
\begin{tabular}{llrrr|rr|rr|rrr}
&&&\multicolumn{2}{c|}{vocabulary}&\multicolumn{2}{c|}{{\sc Bleu}} & \multicolumn{2}{c|}{{\sc chrF3}} & \multicolumn{3}{c}{unigram F$_1$ (\%)}\\
name & segmentation & shortlist & source & target & single & ens-8 & single & ens-8 & all & rare & OOV\\
\hline
\multicolumn{3}{l}{phrase-based \cite{haddow-EtAl:2015:WMT}} & & & 
24.3 & - & 
53.8 & - & 56.0 & 31.3 & 16.5\\ 
\hline
WUnk & - & - & \num{300000} & \num{500000} & 
18.8 & 22.4 & 
46.5 & 49.9 & 54.2 & 25.2 & 0.0\\ 
WDict & - & - & \num{300000} & \num{500000} & 
19.1 & 22.8 & 
47.5 & 51.0 & 54.8 & 26.5 & 6.6\\ 
C2-50k & char-bigram & \num{50000} & \num{60000} & \num{60000} & 
\textbf{20.9} & \textbf{24.1} &
49.0 & 51.6 & 55.2 & 27.8 & 17.4\\ 
BPE-60k & BPE & - & \num{60000} & \num{60000} &
20.5 & 23.6 &
\textbf{49.8} & 52.7 & 55.3 & 29.7 & 15.6 \\ 
BPE-J90k & BPE (joint)& - & \num{90000} & \num{100000} &
20.4 & \textbf{24.1} &
49.7 & \textbf{53.0} & \textbf{55.8} & \textbf{29.7} & \textbf{18.3}\\ 
\end{tabular}
\caption{English$\to$Russian translation performance ({\sc Bleu}, {\sc chrF3} and unigram F$_1$) on newstest2015.
Ens-8: ensemble of 8 models.
Best NMT system in bold.
Unigram F$_1$ (with ensembles) is computed for all words ($n=55654$), rare words (not among top \num{50000} in training set; $n=5442$), and OOVs (not in training set; $n=851$).}
\label{results-russian}
\end{table*}

As a further comment on our translation results, we want to emphasize that performance variability is still an open problem with NMT.
On our development set, we observe differences of up to 1 {\sc Bleu} between different models.
For single systems, we report the results of the model that performs best on dev (out of 8), which has a stabilizing effect, but how to control for randomness deserves further attention in future research.

\section{Analysis}
\label{sec-analysis}

\subsection{Unigram accuracy}

Our main claims are that the translation of rare and unknown words is poor in word-level NMT models, and that subword models improve the translation of these word types.
To further illustrate the effect of different subword segmentations on the translation of rare and unseen words, we plot target-side words sorted by their frequency in the training set.\footnote{We perform binning of words with the same training set frequency, and apply bezier smoothing to the graph.}
To analyze the effect of vocabulary size, we also include the system \textbf{C2-3/500k}, which is a system with the same vocabulary size as the WDict baseline, and character bigrams to represent unseen words.

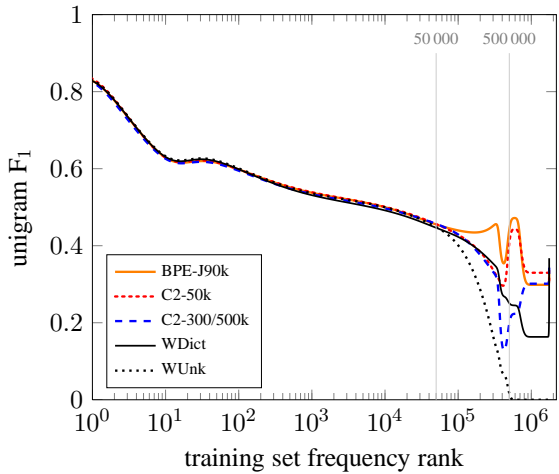
\begin{figure}

\begin{tikzpicture}[scale=0.9]
\pgfplotsset{major grid style={style=dotted,color=black!20}}
\begin{semilogxaxis}[xlabel=training set frequency rank,
    xmin=1,
    xmax=2200000,
    ymin=0,
    ymax=1,
    ylabel=unigram F$_1$,
    legend pos = south west,
    legend style={
        font=\scriptsize,
        /tikz/nodes={anchor=west}
        },
    mark size = 0.1,
    ]

    \addplot +[orange, no markers, raw gnuplot, solid, line width=0.2ex, id=de-bpej] gnuplot {set samples 2000,2000; plot 'data/german-33+36.plot' smooth sbezier;};
    \addplot +[red, no markers, raw gnuplot, dotted, line cap=round, line width=0.2ex, id=de-50k] gnuplot {set samples 2000,2000; plot 'data/german-16+39.plot' smooth sbezier;};
    \addplot +[blue, no markers, raw gnuplot, dashed, line width=0.2ex, id=de-500k] gnuplot {set samples 2000,2000; plot 'data/german-14+38.plot' smooth sbezier;};
    \addplot +[black, no markers, raw gnuplot, line width=0.15ex, id=de-500kdict] gnuplot {set samples 2000,2000; plot 'data/german-mapped.19+35.plot' smooth sbezier;};
    \addplot +[black, no markers, raw gnuplot, dotted, line width=0.2ex, id=de-500knone] gnuplot {set samples 2000,2000; plot 'data/german-19+35.plot' smooth sbezier;};

    \addlegendentry{BPE-J90k}
    \addlegendentry{C2-50k}
    \addlegendentry{C2-300/500k}
    \addlegendentry{WDict}
    \addlegendentry{WUnk}

    \draw[black!20] (axis cs:50000,0) -- (axis cs:50000,0.9) node [gray, above] {\scriptsize{\num{50000}}};
    \draw[black!20] (axis cs:500000,0) -- (axis cs:500000,0.9) node [gray, above] {\scriptsize{\num{500000}}};

\end{semilogxaxis}
\end{tikzpicture} 
\caption{English$\to$German unigram F$_1$ on newstest2015 plotted by training set frequency rank for different NMT systems.}
\label{f1-figure}
\end{figure}

Figure \ref{f1-figure} shows results for the English--German ensemble systems on newstest2015.
Unigram F$_1$ of all systems tends to decrease for lower-frequency words.
The baseline system has a spike in F$_1$ for OOVs, i.e.\ words that do not occur in the training text.
This is because a high proportion of OOVs are names, for which a copy from the source to the target text is a good strategy for English$\to$German.

The systems with a target vocabulary of \num{500000} words mostly differ in how well they translate words with rank > \num{500000}.
A back-off dictionary is an obvious improvement over producing UNK, but the subword system C2-3/500k achieves better performance.
Note that all OOVs that the back-off dictionary produces are words that are copied from the source, usually names, while the subword systems can productively form new words such as compounds.

For the \num{50000} most frequent words, the representation is the same for all neural networks, and all neural networks achieve comparable unigram F$_1$ for this category.
For the interval between frequency rank \num{50000} and \num{500000}, the comparison between C2-3/500k and C2-50k unveils an interesting difference.
The two systems only differ in the size of the shortlist, with C2-3/500k representing words in this interval as single units, and C2-50k via subword units.
We find that the performance of C2-3/500k degrades heavily up to frequency rank \num{500000}, at which point the model switches to a subword representation and performance recovers.
The performance of C2-50k remains more stable.
We attribute this to the fact that subword units are less sparse than words.
In our training set, the frequency rank \num{50000} corresponds to a frequency of 60 in the training data; the frequency rank \num{500000} to a frequency of 2.
Because subword representations are less sparse, reducing the size of the network vocabulary, and representing more words via subword units, can lead to better performance.

The F$_1$ numbers hide some qualitative differences between systems.
For English$\to$German, WDict produces few OOVs (26.5\% recall), but with high precision (60.6\%) , whereas the subword systems achieve higher recall, but lower precision.
We note that the character bigram model C2-50k produces the most OOV words, and achieves relatively low precision of 29.1\% for this category.
However, it outperforms the back-off dictionary in recall (33.0\%).
BPE-60k, which suffers from transliteration (or copy) errors due to segmentation inconsistencies, obtains a slightly better precision (32.4\%), but a worse recall (26.6\%).
In contrast to BPE-60k, the joint BPE encoding of BPE-J90k improves both precision (38.6\%) and recall (29.8\%).

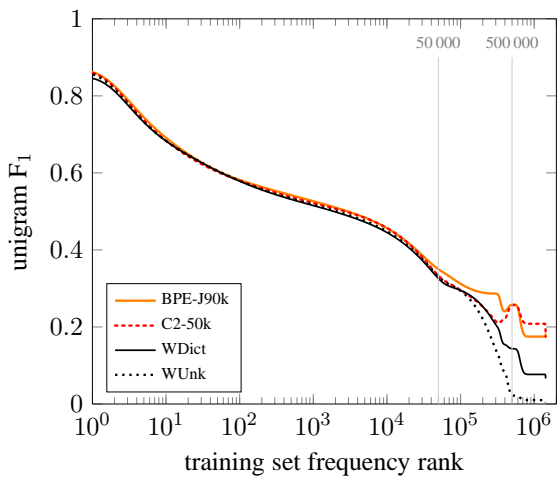
\begin{figure}

\begin{tikzpicture}[scale=0.9]
\pgfplotsset{major grid style={style=dotted,color=black!20}}
\begin{semilogxaxis}[xlabel=training set frequency rank,
    xmin=1,
    xmax=2000000,
    ymin=0,
    ymax=1,
    ylabel=unigram F$_1$,
    legend pos = south west,
    legend style={
        font=\scriptsize,
        /tikz/nodes={anchor=west}
        },
    mark size = 0.1,
    ]

    \addplot +[orange, no markers, raw gnuplot, solid, line width=0.2ex, id=ru-bpej] gnuplot {set samples 2000,2000; plot 'data/russian-6+8.plot' smooth sbezier;};
    \addplot +[red, no markers, raw gnuplot, dotted, line cap=round, line width=0.2ex, id=ru-50k] gnuplot {set samples 2000,2000; plot 'data/russian-3+10.plot' smooth sbezier;};
    \addplot +[black, no markers, raw gnuplot, line width=0.15ex, id=ru-500kdict] gnuplot {set samples 2000,2000; plot 'data/russian-mapped.1+9.plot' smooth sbezier;};
    \addplot +[black, no markers, raw gnuplot, dotted, line width=0.2ex, id=ru-500knone] gnuplot {set samples 2000,2000; plot 'data/russian-1+9.plot' smooth sbezier;};

    \addlegendentry{BPE-J90k}
    \addlegendentry{C2-50k}
    \addlegendentry{WDict}
    \addlegendentry{WUnk}

    \draw[black!20] (axis cs:50000,0) -- (axis cs:50000,0.9) node [gray, above] {\scriptsize{\num{50000}}};
    \draw[black!20] (axis cs:500000,0) -- (axis cs:500000,0.9) node [gray, above] {\scriptsize{\num{500000}}};

\end{semilogxaxis}
\end{tikzpicture} 
\caption{English$\to$Russian unigram F$_1$ on newstest2015 plotted by training set frequency rank for different NMT systems.}
\label{f1-figure2}
\end{figure}

For English$\to$Russian, unknown names can only rarely be copied, and usually require transliteration.
Consequently, the WDict baseline performs more poorly for OOVs (9.2\% precision; 5.2\% recall), and the subword models improve both precision and recall (21.9\% precision and 15.6\% recall for BPE-J90k).
The full unigram F$_1$ plot is shown in Figure \ref{f1-figure2}.

\subsection{Manual Analysis}

Table \ref{examples-de} shows two translation examples for the translation direction English$\to$German,
Table \ref{examples-ru} for English$\to$Russian.
The baseline system fails for all of the examples, either by deleting content (\emph{health}), or by copying source words that should be translated or transliterated.
The subword translations of \emph{health research institutes} show that the subword systems are capable of learning translations when oversplitting (\emph{research}$\rightarrow$\emph{Fo\pipe rs\pipe ch\pipe un\pipe g}),
or when the segmentation does not match morpheme boundaries: the segmentation \emph{Forschungs\pipe instituten} would be linguistically more plausible, and simpler to align to the English \emph{research institutes},
than the segmentation \emph{Forsch\pipe ungsinstitu\pipe ten} in the BPE-60k system, but still, a correct translation is produced.
If the systems have failed to learn a translation due to data sparseness, like for \emph{asinine}, which should be translated as \emph{dumm}, we see translations that are wrong, but could be plausible for (partial) loanwords (\emph{asinine Situation}$\rightarrow$\emph{Asinin-Situation}).

The English$\to$Russian examples show that the subword systems are capable of transliteration.
However, transliteration errors do occur, either due to ambiguous transliterations, or because of non-consistent segmentations between source and target text which make it hard for the system to learn a transliteration mapping.
Note that the BPE-60k system encodes \emph{Mirzayeva} inconsistently for the two language pairs (\emph{Mirz\pipe ayeva}$\rightarrow$\foreignlanguage{russian}{Мир\pipe за\pipe ева} \emph{Mir\pipe za\pipe eva}).
This example is still translated correctly, but we observe spurious insertions and deletions of characters in the BPE-60k system.
An example is the transliteration of \emph{rakfisk}, where a \emph{\foreignlanguage{russian}{п}} is inserted and a \emph{\foreignlanguage{russian}{к}} is deleted.
We trace this error back to translation pairs in the training data with inconsistent segmentations, such as (\emph{p\pipe rak\pipe ri\pipe ti}$\rightarrow$\emph{\foreignlanguage{russian}{пра\pipe крит\pipe и}} (pra\pipe krit\pipe i)), from which the translation (\emph{rak}$\rightarrow$\emph{\foreignlanguage{russian}{пра}}) is erroneously learned.
The segmentation of the joint BPE system (BPE-J90k) is more consistent (\emph{pra\pipe krit\pipe i}$\rightarrow$\emph{\foreignlanguage{russian}{пра\pipe крит\pipe и}} (pra\pipe krit\pipe i)).

\begin{table}
\centering
\small
\begin{tabular}{l|l}
system & sentence\\
\hline
source & health research institutes \\ 
reference & Gesundheitsforschungsinstitute \\
WDict & Forschungsinstitute \\
C2-50k & Fo\pipe rs\pipe ch\pipe un\pipe gs\pipe in\pipe st\pipe it\pipe ut\pipe io\pipe ne\pipe n\\
BPE-60k & Gesundheits\pipe forsch\pipe ungsinstitu\pipe ten \\
BPE-J90k & Gesundheits\pipe forsch\pipe ungsin\pipe stitute \\
\hline
source & asinine situation \\ 
reference & dumme Situation \\
WDict & asinine situation $\rightarrow$ UNK $\rightarrow$ asinine \\
C2-50k & as\pipe in\pipe in\pipe e situation $\rightarrow$ As\pipe in\pipe en\pipe si\pipe tu\pipe at\pipe io\pipe n\\
BPE-60k & as\pipe in\pipe ine situation $\rightarrow$ A\pipe in\pipe line-\pipe Situation \\
BPE-J90K & as\pipe in\pipe ine situation $\rightarrow$ As\pipe in\pipe in-\pipe Situation \\
\end{tabular}
\caption{English$\to$German translation example. ``\pipe '' marks subword boundaries.}
\label{examples-de}
\end{table}

\begin{table}
\centering
\small
\begin{tabular}{l|l}
system & sentence\\
\hline
source & Mirzayeva \\ 
reference & \foreignlanguage{russian}{Мирзаева} (Mirzaeva)\\
WDict & Mirzayeva\phantom{\pipe\pipe\pipe\pipe} $\rightarrow$ UNK $\rightarrow$ Mirzayeva \\
C2-50k & Mi\pipe rz\pipe ay\pipe ev\pipe a $\rightarrow$ \foreignlanguage{russian}{Ми\pipe рз\pipe ае\pipe ва} (Mi\pipe rz\pipe ae\pipe va)\\
BPE-60k & Mirz\pipe ayeva\phantom{\pipe\pipe\pipe} $\rightarrow$ \foreignlanguage{russian}{Мир\pipe за\pipe ева} (Mir\pipe za\pipe eva)\\
BPE-J90k & Mir\pipe za\pipe yeva\phantom{\pipe\pipe} $\rightarrow$ \foreignlanguage{russian}{Мир\pipe за\pipe ева} (Mir\pipe za\pipe eva)\\
\hline
source & rakfisk \\ 
reference & \foreignlanguage{russian}{ракфиска} (rakfiska)\\
WDict & rakfisk\phantom{\pipe\pipe\pipe} $\rightarrow$ UNK $\rightarrow$ rakfisk\\
C2-50k & ra\pipe kf\pipe is\pipe k $\rightarrow$ \foreignlanguage{russian}{ра\pipe кф\pipe ис\pipe к} (ra\pipe kf\pipe is\pipe k)\\
BPE-60k & rak\pipe f\pipe isk\phantom{\pipe} $\rightarrow$ \foreignlanguage{russian}{пра\pipe ф\pipe иск} (pra\pipe f\pipe isk)\\
BPE-J90k & rak\pipe f\pipe isk\phantom{\pipe} $\rightarrow$ \foreignlanguage{russian}{рак\pipe ф\pipe иска} (rak\pipe f\pipe iska) \\
\end{tabular}
\caption{English$\to$Russian translation examples. ``\pipe '' marks subword boundaries.}
\label{examples-ru}
\end{table}

\section{Conclusion}

The main contribution of this paper is that we show that neural machine translation systems are capable of open-vocabulary translation by representing rare and unseen words as a sequence of subword units.\footnote{The source code of the segmentation algorithms is available at \url{https://github.com/rsennrich/subword-nmt}.}
This is both simpler and more effective than using a back-off translation model.
We introduce a variant of byte pair encoding for word segmentation, which is capable of encoding open vocabularies with a compact symbol vocabulary of variable-length subword units.
We show performance gains over the baseline with both BPE segmentation, and a simple character bigram segmentation.

Our analysis shows that not only out-of-vocabulary words, but also rare in-vocabulary words are translated poorly by our baseline NMT system,
and that reducing the vocabulary size of subword models can actually improve performance.
In this work, our choice of vocabulary size is somewhat arbitrary, and mainly motivated by comparison to prior work.
One avenue of future research is to learn the optimal vocabulary size for a translation task, which we expect to depend on the language pair and amount of training data, automatically.
We also believe there is further potential in bilingually informed segmentation algorithms to create more alignable subword units, although the segmentation algorithm cannot rely on the target text at runtime.

While the relative effectiveness will depend on language-specific factors such as vocabulary size, we believe that subword segmentations are suitable for most language pairs, eliminating the need for large NMT vocabularies or back-off models.

\section*{Acknowledgments}

We thank Maja Popović for her implementation of {\sc chrF}, with which we verified our re-implementation.
The research presented in this publication was conducted in cooperation with Samsung Electronics Polska sp.\ z o.o.\ - Samsung R\&D Institute Poland.
This project received funding from the European Union's Horizon 2020 research and innovation programme under grant agreement 645452 (QT21).

\bibliographystyle{acl2016}
\bibliography{../bibliography}

\end{document}